\documentclass[11pt,a4paper]{article}
\usepackage[hyperref]{acl2019}
\usepackage{times}
\usepackage{latexsym}
\usepackage{booktabs}
\usepackage{multirow}
\usepackage{url}
\usepackage[para,online,flushleft]{threeparttable}
\usepackage{tabularx}
\usepackage{float}
\usepackage{graphicx}

\aclfinalcopy 
\def\aclpaperid{37} 

\title{Transfer Learning in Biomedical Natural Language Processing: 
An Evaluation of BERT and ELMo on Ten Benchmarking Datasets}

\author{Yifan Peng \hspace{1em} Shankai Yan \hspace{1em} Zhiyong Lu \\
  National Center for Biotechnology Information \\
  National Library of Medicine, National Institutes of Health \\
  Bethesda, MD, USA \\
  \texttt{\{yifan.peng, shankai.yan, zhiyong.lu\}@nih.gov}\\}

\date{}

\begin{document}
\maketitle
\begin{abstract}
Inspired by the success of the General Language Understanding Evaluation benchmark, we introduce the Biomedical Language Understanding Evaluation (BLUE) benchmark to facilitate research in the development of pre-training language representations in the biomedicine domain. The benchmark consists of five tasks with ten datasets that cover both biomedical and clinical texts with different dataset sizes and difficulties. We also evaluate several baselines based on BERT and ELMo and find that the BERT model pre-trained on PubMed abstracts and MIMIC-III clinical notes achieves the best results. We make the datasets, pre-trained models, and codes publicly available at \url{https://github.com/ncbi-nlp/BLUE_Benchmark}.
\end{abstract}

\section{Introduction}

With the growing amount of biomedical information available in textual form, there have been significant advances in the development of pre-training language representations that can be applied to a range of different tasks in the biomedical domain, such as pre-trained word embeddings, sentence embeddings, and contextual representations~\cite{chiu2016how,chen2018biosentvec,peters2017semi,lee2019biobert,smalheiser2019unsupervised}.

In the general domain, we have recently observed that the General Language Understanding Evaluation (GLUE) benchmark~\cite{wang2018glue} has been successfully promoting the development of language representations of general purpose~\cite{peters2017semi,radford2018improving,devlin2018bert}. To the best of our knowledge, however, there is no publicly available benchmarking in the biomedicine domain.

To facilitate research on language representations in the biomedicine domain, we present the Biomedical Language Understanding Evaluation (BLUE) benchmark, which consists of five different biomedicine text-mining tasks with ten corpora. Here, we rely on preexisting datasets because they have been widely used by the BioNLP community as shared tasks~\cite{huang2015community}. These tasks cover a diverse range of text genres (biomedical literature and clinical notes), dataset sizes, and degrees of difficulty and, more importantly, highlight common biomedicine text-mining challenges. We expect that the models that perform better on all or most tasks in BLUE will address other biomedicine tasks more robustly.  

To better understand the challenge posed by BLUE, we conduct experiments with two baselines: One makes use of the BERT model~\cite{devlin2018bert} and one makes use of ELMo~\cite{peters2017semi}. Both are state-of-the-art language representation models and demonstrate promising results in NLP tasks of general purpose. We find that the BERT model pre-trained on PubMed abstracts~\cite{fiorini2018how} and MIMIC-III clinical notes~\cite{johnson2016mimic} achieves the best results, and is significantly superior to other models in the clinical domain. This demonstrates the importance of pre-training among different text genres.

In summary, we offer: (i) five tasks with ten biomedical and clinical text-mining corpora with different sizes and levels of difficulty, (ii) codes for data construction and model evaluation for fair comparisons, (iii) pretrained BERT models on PubMed abstracts and MIMIC-III, and (iv) baseline results.

\section{Related work}

There is a long history of using shared language representations to capture text semantics in biomedical text and data mining research. Such research utilizes a technique, termed transfer learning, whereby the language representations are pre-trained on large corpora and fine-tuned in a variety of downstream tasks, such as named entity recognition and relation extraction.

One established trend is a form of word embeddings that represent the semantic, using high dimensional vectors~\cite{chiu2016how,wang2018comparison,zhang2019improving}. Similar methods also have been derived to improve embeddings of word sequences by introducing sentence embeddings~\cite{chen2018biosentvec}. They always, however, require complicated neural networks to be effectively used in downstream applications.

Another popular trend, especially in recent years, is the context-dependent representation. Different from word embeddings, it allows the meaning of a word to change according to the context in which it is used~\cite{melamud2016role,peters2017semi,devlin2018bert,dai2019transformer}. In the scientific domain, \citet{beltagy2019scibert} released SciBERT which is trained on scientific text. In the biomedical domain, BioBERT~\cite{lee2019biobert} and BioELMo~\cite{jin2019probing} were pre-trained and applied to several specific tasks. In the clinical domain, \citet{alsentzer2019publicly} released a clinical BERT base model trained on the MIMIC-III database. Most of these works, however, were evaluated on either different datasets or the same dataset with slightly different sizes of examples. This makes it challenging to fairly compare various language models. 

Based on these reasons, a standard benchmarking is urgently required. Parallel to our work, \citet{lee2019biobert} introduced three tasks: named entity recognition, relation extraction, and QA, while \citet{jin2019probing} introduced NLI in addition to named entity recognition. To this end, we deem that BLUE is different in three ways. First, BLUE is selected to cover a diverse range of text genres, including both biomedical and clinical domains.  Second, BLUE goes beyond sentence or sentence pairs by including document classification tasks. Third, BLUE provides a comprehensive suite of codes to reconstruct dataset from scratch without removing any instances.

\section{Tasks}

BLUE contains five tasks with ten corpora that cover a broad range of data quantities and difficulties (Table~\ref{tab:tasks}). Here, we rely on preexisting datasets because they have been widely used by the BioNLP community as shared tasks.

\begin{table*}[!ht]
\centering
\small
\begin{tabular}{lrrrlllr}
\toprule
Corpus & Train & Dev & Test & Task & Metrics & Domain & Avg \\
&&&&&&&sent len\\
\midrule
MedSTS, sentence pairs & 675 & 75 & 318 & Sentence similarity & Pearson  & Clinical & 25.8\\
BIOSSES, sentence pairs & 64 & 16 & 20 & Sentence similarity & Pearson  & Biomedical & 22.9\\
BC5CDR-disease, mentions & 4182 & 4244 & 4424 & NER & F1 & Biomedical & 22.3\\
BC5CDR-chemical, mentions & 5203 & 5347 & 5385 & NER & F1 & Biomedical & 22.3\\
ShARe/CLEFE, mentions & 4628 & 1075 & 5195  & NER & F1 & Clinical & 10.6\\
DDI, relations & 2937 & 1004 & 979 & Relation extraction & micro F1 & Biomedical & 41.7\\
ChemProt, relations & 4154  & 2416 & 3458 & Relation extraction & micro F1 & Biomedical & 34.3\\
i2b2 2010, relations & 3110 & 11 & 6293 & Relation extraction & F1 & Clinical & 24.8\\
HoC, documents & 1108 & 157 & 315 & Document classification & F1 & Biomedical & 25.3\\
MedNLI, pairs & 11232 & 1395 & 1422 & Inference & accuracy & Clinical & 11.9\\
\bottomrule
\end{tabular}
\caption{BLUE tasks}
\label{tab:tasks}
\end{table*}

\begin{figure}[t]
\centering
\includegraphics[height=1ex]{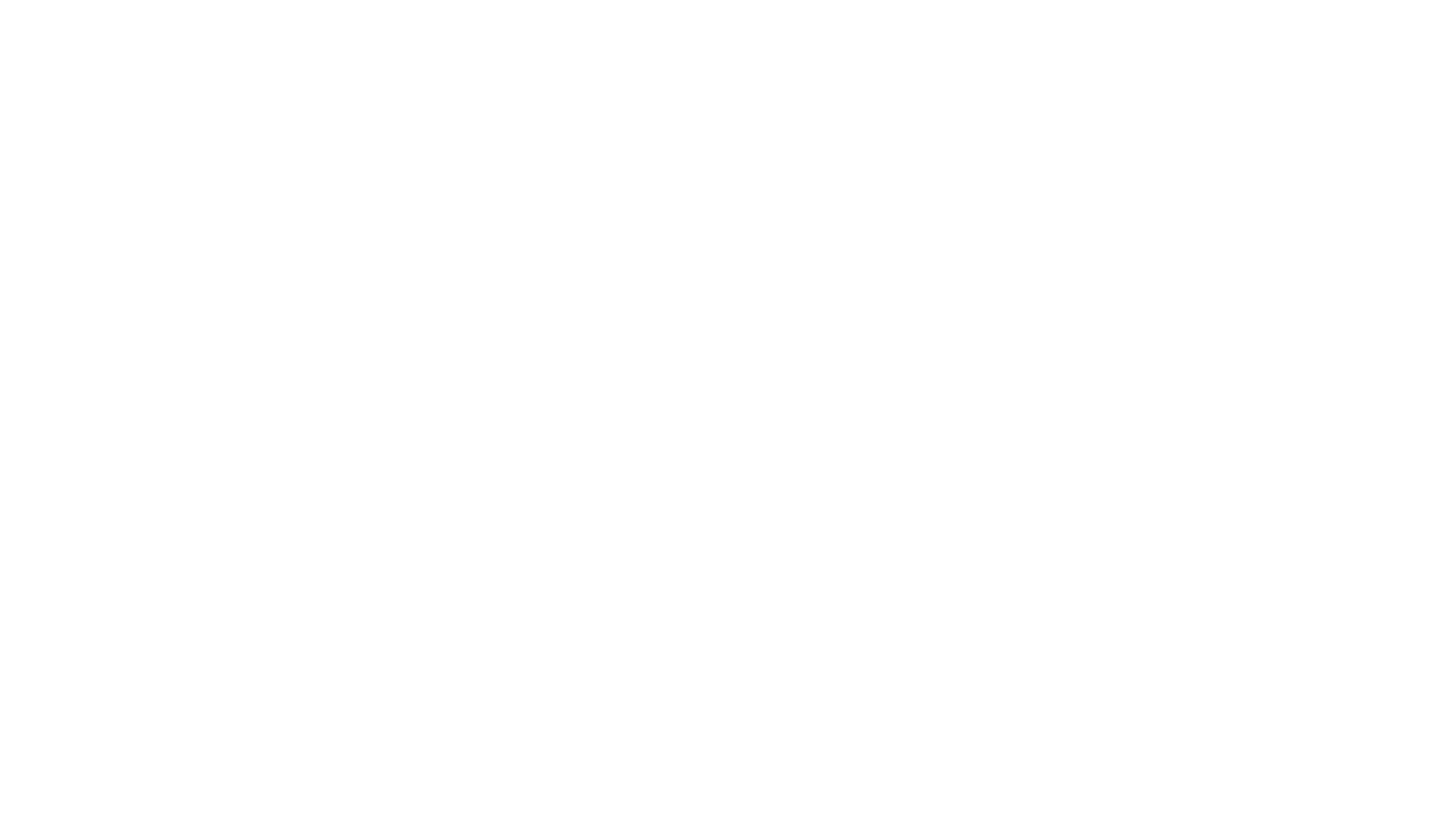}
\label{fig:example}
\vspace{-5ex}
\end{figure}

\subsection{Sentence similarity}

The sentence similarity task is to predict similarity scores based on sentence pairs. Following common practice, we evaluate similarity by using Pearson correlation coefficients.

\textbf{BIOSSES} is a corpus of sentence pairs selected from the Biomedical Summarization Track Training Dataset in the biomedical domain~\cite{sogancioglu2017biosses}.\footnote{\url{http://tabilab.cmpe.boun.edu.tr/BIOSSES/}} To develop BIOSSES, five curators judged their similarity, using scores that ranged from 0 (no relation) to 4 (equivalent). Here, we randomly select 80\% for training and 20\% for testing because there is no standard splits in the released data.

\textbf{MedSTS} is a corpus of sentence pairs selected from Mayo Clinic’s clinical data warehouse~\cite{wang2018medstsa}. To develop MedSTS, two medical experts graded the sentence's semantic similarity scores from 0 to 5 (low to high similarity). We use the standard training and testing sets in the shared task.

\subsection{Named entity recognition}

The aim of the named entity recognition task is to predict mention spans given in the text~\cite{jurafsky2008speech}. The results are evaluated through a comparison of the set of mention spans annotated within the document with the set of mention spans predicted by the model. We evaluate the results by using the strict version of precision, recall, and F1-score. For disjoint mentions, all spans also must be strictly correct. To construct the dataset, we used spaCy\footnote{\url{https://spacy.io/}} to split the text into a sequence of tokens when the original datasets do not provide such information.

\textbf{BC5CDR} is a collection of 1,500 PubMed titles and abstracts selected from the CTD-Pfizer corpus and was used in the BioCreative V chemical-disease relation task~\cite{li2016biocreative}.\footnote{\url{https://biocreative.bioinformatics.udel.edu/tasks/biocreative-v/track-3-cdr/}} The diseases and chemicals mentioned in the articles were annotated independently by two human experts with medical training and curation experience. We use the standard training and test set in the BC5CDR shared task~\cite{wei2016assessing}.

\textbf{ShARe/CLEF} eHealth Task 1 Corpus is a collection of 299 deidentified clinical free-text notes from the MIMIC II database~\cite{suominen2013overview}.\footnote{\url{https://physionet.org/works/ShAReCLEFeHealth2013/}} The disorders mentioned in the clinical notes were annotated by two professionally trained annotators, followed by an adjudication step, resulting in high inter-annotator agreement. We use the standard training and test set in the ShARe/CLEF eHealth Tasks 1.

\subsection{Relation extraction}

The aim of the relation extraction task is to predict relations and their types between the two entities mentioned in the sentences. The relations with types were compared to annotated data. We use the standard micro-average precision, recall, and F1-score metrics.

\textbf{DDI} extraction 2013 corpus is a collection of 792 texts selected from the DrugBank database and other 233 Medline abstracts~\cite{herrero-zazo2013ddi}.\footnote{\url{http://labda.inf.uc3m.es/ddicorpus}} The drug-drug interactions, including both pharmacokinetic and pharmacodynamic interactions, were annotated by two expert pharmacists with a substantial background in pharmacovigilance. In our benchmark, we use 624 train files and 191 test files to evaluate the performance and report the micro-average F1-score of the four DDI types.

\textbf{ChemProt} consists of 1,820 PubMed abstracts with chemical-protein interactions annotated by domain experts and was used in the BioCreative VI text mining chemical-protein interactions shared task~\cite{krallinger2017overview}.\footnote{\url{https://biocreative.bioinformatics.udel.edu/news/corpora/chemprot-corpus-biocreative-vi/}} We use the standard training and test sets in the ChemProt shared task and evaluate the same five classes: CPR:3, CPR:4, CPR:5, CPR:6, and CPR:9.

\textbf{i2b2 2010} shared task collection consists of 170 documents for training and 256 documents for testing, which is the subset of the original dataset~\cite{uzuner20112010}.\footnote{\url{https://www.i2b2.org/NLP/DataSets/}} The dataset was collected from three different hospitals and was annotated by medical practitioners for eight types of relations between problems and treatments.

\subsection{Document multilabel classification}

The multilabel classification task predicts multiple labels from the texts. 

\textbf{HoC} (the Hallmarks of Cancers corpus) consists of 1,580 PubMed abstracts annotated with ten currently known hallmarks of cancer~\cite{baker2016automatic}.\footnote{\url{https://www.cl.cam.ac.uk/~sb895/HoC.html}} Annotation was performed at sentence level by an expert with 15+ years of experience in cancer research. We use 315 ($\sim $20\%) abstracts for testing and the remaining abstracts for training. For the HoC task, we followed the common practice and reported the example-based F1-score on the abstract level~\cite{zhang2014review,du2018ml}.

\subsection{Inference task}

The aim of the inference task is to predict whether the premise sentence entails or contradicts the hypothesis sentence. We use the standard overall accuracy to evaluate the performance.

\textbf{MedNLI} is a collection of sentence pairs selected from MIMIC-III~\cite{romanov2018lessons}.\footnote{\url{https://physionet.org/physiotools/mimic-code/mednli/}} Given a premise sentence and a hypothesis sentence, two board-certified radiologists graded whether the task predicted whether the premise entails the hypothesis (entailment), contradicts the hypothesis (contradiction), or neither (neutral). We use the same training, development, and test sets in Romanov and Shivade~\cite{romanov2018lessons}.

\subsection{Total score}

Following the practice in~\citet{wang2018glue} and \citet{lee2019biobert}, we use a macro-average of F1-scores and Pearson scores to determine a system's position.

\section{Baselines}

For baselines, we evaluate several pre-training models as described below. The original code for the baselines is available at \url{https://github.com/ncbi-nlp/NCBI_BERT}.

\subsection{BERT}

\subsubsection{Pre-training BERT}

BERT~\cite{devlin2018bert} is a contextualized word representation model that is pre-trained based on a masked language model, using bidirectional Transformers~\cite{vaswani2017attentiona}.

In this paper, we pre-trained our own model BERT on PubMed abstracts and clinical notes (MIMIC-III). The statistics of the text corpora on which BERT was pre-trained are shown in Table~\ref{tab:corpora}.

\begin{table}[H]
\centering
\begin{tabular}{lrl}
\toprule
Corpus & Words & Domain\\
\midrule
PubMed abstract & $>$ 4,000M & Biomedical\\
MIMIC-III & $>$ 500M & Clinical\\
\bottomrule
\end{tabular}
\caption{Corpora}
\label{tab:corpora}
\end{table}

We initialized BERT with pre-trained BERT provided by~\cite{devlin2018bert}. We then continue to pre-train the model, using the listed corpora. 

We released our BERT-Base and BERT-Large models, using the same vocabulary, sequence length, and other configurations provided by~\citet{devlin2018bert}. Both models were trained with 5M steps on the PubMed corpus and 0.2M steps on the MIMIC-III corpus.

\subsubsection{Fine-tuning with BERT}

BERT is applied to various downstream text-mining tasks while requiring only minimal architecture modification. 

For sentence similarity tasks, we packed the sentence pairs together into a single sequence, as suggested in~\citet{devlin2018bert}.

For named entity recognition, we used the BIO tags for each token in the sentence. We considered the tasks similar to machine translation, as predicting the sequence of BIO tags from the input sentence. 

We treated the relation extraction task as a sentence classification by replacing two named entity mentions of interest in the sentence with pre-defined tags (e.g., @GENE\$, @DRUG\$)~\cite{lee2019biobert}. For example, we used ``@CHEMICAL\$ protected against the RTI-76-induced inhibition of @GENE\$ binding.'' to replace the original sentence ``Citalopram protected against the RTI-76-induced inhibition of SERT binding.'' in which ``citalopram'' and ``SERT'' has a chemical-gene relation.

For multi-label tasks, we fine-tuned the model to predict multi-labels for each sentence in the document. We then combine the labels in one document and compare them with the gold-standard.

Like BERT, we provided sources code for fine-tuning, prediction, and evaluation to make it straightforward to follow those examples to use our BERT pre-trained models for all tasks.

\subsection{Fine-tuning with ELMo}

\begin{table*}[!ht]
\centering
\begin{threeparttable}
\begin{tabularx}{\textwidth}{Xlccccccc}
\toprule
\multirow{3}{*}{Task} & \multirow{3}{*}{Metrics} &  \multirow{3}{*}{SOTA*} & \multirow{3}{*}{ELMo} & \multirow{3}{*}{BioBERT} & \multicolumn{4}{c}{Our BERT}\\
\cmidrule{6-9}
 &  &  &  &  & Base & Base & Large & Large\\
 &  &  &  &  & (P) & (P+M) & (P) & (P+M) \\
\midrule
MedSTS & Pearson & 83.6 & 68.6 & 84.5 & 84.5 & \textbf{84.8} & 84.6 & 83.2\\
BIOSSES & Pearson & 84.8 & 60.2 & 82.7 & 89.3 & \textbf{91.6} & 86.3 & 75.1\\
BC5CDR-disease & F & 84.1 & 83.9 & 85.9 & \textbf{86.6} & 85.4 & 82.9 & 83.8\\
BC5CDR-chemical & F & 93.3 & 91.5 & 93.0 & \textbf{93.5} & 92.4 & 91.7 & 91.1\\
ShARe/CLEFE & F & 70.0 & 75.6 & 72.8 & 75.4 & \textbf{77.1} & 72.7 & 74.4\\
DDI & F & 72.9 & 78.9 & 78.8 & 78.1 & 79.4 & \textbf{79.9} & 76.3\\
ChemProt & F & 64.1 & 66.6 & 71.3 & 72.5 & 69.2 & \textbf{74.4} & 65.1\\
i2b2 & F & 73.7 & 71.2 & 72.2 & 74.4 & \textbf{76.4} & 73.3 & 73.9\\
HoC & F & 81.5 & 80.0 & 82.9 & 85.3 & 83.1 & \textbf{87.3} & 85.3\\
MedNLI & acc & 73.5 & 71.4 & 80.5 & 82.2 & \textbf{84.0} & 81.5 & 83.8\\
\midrule
Total &  &  & 78.8 &  80.5 & 82.2 & \textbf{82.3} & 81.5 & 79.2\\
\bottomrule
\end{tabularx}
\begin{tablenotes}
\item[*] SOTA, state-of-the-art as of April 2019, to the best of our knowledge: MedSTS, BIOSSES~\cite{chen2018biosentvec}; BC5CDR-disease, BC5CDR-chem~\cite{yoon2018collabonet}; ShARe/CLEFE~\cite{leaman2015challengesa}; DDI~\cite{zhang2018drug}. Chem-Prot~\cite{peng2018extracting}; i2b2~\cite{rink2011automatic}; HoC~\cite{du2018ml}; MedNLI~\cite{romanov2018lessons}. P: PubMed, P+M: PubMed + MIMIC-III
\end{tablenotes}
\end{threeparttable}
\caption{Baseline performance on the BLUE task test sets.}
\label{tab:results}
\end{table*}

We adopted the ELMo model pre-trained on PubMed abstracts~\cite{peters2017semi} to accomplish the BLUE tasks.\footnote{\url{https://allennlp.org/elmo}} The output of ELMo embeddings of each token is used as input for the fine-tuning model. We retrieved the output states of both layers in ELMo and concatenated them into one vector for each word. We used the maximum sequence length 128 for padding. The learning rate was set to 0.001 with an Adam optimizer. We iterated the training process for 20 epochs with batch size 64 and early stopped if the training loss did not decrease.

For sentence similarity tasks, we used bag of embeddings with the average strategy to transform the sequence of word embeddings into a sentence embedding. Afterward, we concatenated two sentence embeddings and fed them into an architecture with one dense layer to predict the similarity of two sentences.

For named entity recognition, we used a Bi-LSTM-CRF implementation as a sequence tagger~\cite{huang2015bidirectional,si2019enhancing,lample2016neural}. Specifically, we concatenated the GloVe word embeddings~\cite{pennington2014glove}, character embeddings, and ELMo embeddings of each token and fed the combined vectors into the sequence tagger to predict the label for each token. The GloVe word embeddings\footnote{\url{https://nlp.stanford.edu/projects/glove/}} and character embeddings have 100 and 25 dimensions, respectively. The hidden sizes of the Bi-LSTM are also set to 100 and 25 for the word and character embeddings, respectively.

For relation extraction and multi-label tasks, we followed the steps in fine-tuning with BERT but used the averaged ELMo embeddings of all words in each sentence as the sentence embedding.

\section{Benchmark results and discussion}

We pre-trained four BERT models: BERT-Base (P), BERT-Large (P), BERT-Base (P+M), BERT-Large (P+M) on PubMed abstracts only, and the combination of PubMed abstracts and clinical notes, respectively. We present performance on the main benchmark tasks in Table~\ref{tab:results}. More detailed comparison is shown in the Appendix~\ref{sec:appendix}.

Overall, our BERT-Base (P+M) that were pre-trained on both PubMed abstract and MIMIC-III achieved the best results across five tasks, even though it is only slightly better than the one pre-trained on PubMed abstracts only. Compared to the tasks in the clinical domain and biomedical domain, BERT-Base (P+M) is significantly superior to other models. This demonstrates the importance of pre-training among different text genres. 

When comparing BERT pre-trained using the base settings against that using the large settings, it is a bit surprising that BERT-Base is better than BERT-Large except in relation extraction and document classification tasks. Further analysis shows that, on these tasks, the average length of sentences is longer than those of others (Table~\ref{tab:tasks}). In addition, BERT-Large pre-trained on PubMed and MIMIC is worse than other models overall. However, BERT-Large (P) performs the best in the multilabel task, even compared with the feature-based model utilizing enriched ontology~\cite{yan2017elucidating}. This is partially because the MIMIC-III data are relatively smaller than the PubMed abstracts and, thus, cannot pre-train the large model sufficiently.

In the sentence similarity tasks, BERT-Base (P+M) achieves the best results on both datasets. Because the BIOSSES dataset is very small (there are only 16 sentence pairs in the test set), all BERT models' performance was unstable. This problem has also been noted in the work of \citet{devlin2018bert} when the model was evaluated on the GLUE benchmarking. Here, we obtained the best results by following the same strategy: selecting the best model on the development set after several runs. Other possible ways to overcome this issue include choosing the model with the best performance from multiple runs or averaging results from multiple fine-tuned models.

In the named entity recognition tasks, BERT-Base (P) achieved the best results on two biomedical datasets, whereas BERT-Base (P+M) achieved the best results on the clinical dataset. In all cases, we observed that the winning model obtained higher recall than did the others. Given that we use the pre-defined vocabulary in the original BERT and that this task relies heavily on the tokenization, it is possible that using BERT as pertaining to a custom sentence piece tokenizer may further improve the model's performance.

\section{Conclusion}

In this study, we introduce BLUE, a collection of resources for evaluating and analyzing biomedical natural language representation models. We find that the BERT models pre-trained on PubMed abstracts and clinical notes see better performance than do most state-of-the-art models. Detailed analysis shows that our benchmarking can be used to evaluate the capacity of the models to understand the biomedicine text and, moreover, to shed light on the future directions for developing biomedicine language representations.

\section*{Acknowledgments}

This work was supported by the Intramural Research Programs of the NIH National Library of Medicine. This work was
supported by the National Library of Medicine of the National Institutes of Health under award number K99LM013001-01. We are also grateful to shared task organizers and the authors of BERT and ELMo to make the data and codes publicly available. 

\bibliography{acl2019}
\bibliographystyle{acl_natbib}

\appendix

\section{Appendices}
\label{sec:appendix}

TP: true positive, FP: false positive, FN: false negative, P: precision, R: recall, F1: F1-score

\subsection{Named Entity Recognition}

\noindent
\small
\begin{tabularx}{.5\textwidth}{Xr@{\hspace{2ex}}r@{\hspace{2ex}}r@{\hspace{2ex}}r@{\hspace{2ex}}r@{\hspace{2ex}}r}
\toprule
BC5CDR-disease & TP & FP & FN & P & R & F1\\
\midrule
\cite{yoon2018collabonet} & - & - & - & 85.6 & 82.6 & 84.1\\
ELMo & 3740 & 749 & 684 & 83.3 & 84.5 & 83.9\\
BioBERT & 3807 & 637 & 617 & 85.7 & 86.1 & 85.9\\
Our BERT &  &  &  &  &  & \\
\hspace{1em}Base (P) & 3806 & 635 & 564 & 85.9 & 87.3 & 86.6\\
\hspace{1em}Base (P+M) & 3788 & 655 & 636 & 85.3 & 85.6 & 85.4\\
\hspace{1em}Large (P)& 3729 & 847 & 695 & 81.5 & 84.3 & 82.9\\
\hspace{1em}Large (P+M) & 3765 & 799 & 659 & 82.5 & 85.1 & 83.8\\
\bottomrule
\end{tabularx}
\vspace{1em}

\noindent
\begin{tabularx}{.5\textwidth}{Xr@{\hspace{2ex}}r@{\hspace{2ex}}r@{\hspace{2ex}}r@{\hspace{2ex}}r@{\hspace{2ex}}r}
\toprule
BC5CDR-chemical & TP & FP & FN & P & R & F1\\
\midrule
\cite{yoon2018collabonet}  & - & - & - & 94.3 & 92.4 & 93.3\\
ELMo & 4864 & 386 & 521 & 92.6 & 90.3 & 91.5\\
BioBERT & 5029 & 404 & 356 & 92.6 & 93.4 & 93.0\\
Our BERT &  &  &  &  &  & \\
\hspace{1em}Base (P) & 5027 & 336 & 358 & 93.7 & 93.4 & 93.5\\
\hspace{1em}Base (P+M) & 4914 & 341 & 471 & 93.5 & 91.3 & 92.4\\
\hspace{1em}Large (P) & 4941 & 454 & 444 & 91.6 & 91.8 & 91.7\\
\hspace{1em}Large (P+M) & 4905 & 484 & 480 & 91.0 & 91.1 & 91.1\\
\bottomrule
\end{tabularx}
\vspace{1em}

\noindent
\begin{tabularx}{.5\textwidth}{X@{\hspace{1ex}}r@{\hspace{1ex}}r@{\hspace{1ex}}r@{\hspace{2ex}}r@{\hspace{2ex}}r@{\hspace{2ex}}r}
\toprule
ShARe/CLEFE & TP & FP & FN & P & R & F1\\
\midrule
\cite{leaman2015challengesa} & - & - & - & 79.7 & 71.3 & 75.3\\
ELMo & 3928 & 1117 & 1423 & 77.9 & 73.4 & 75.6\\
BioBERT & 3898 & 1024 & 1453 & 79.2 & 72.8 & 75.9\\
Our BERT &  &  &  &  &  & \\
\hspace{1em}Base (P) & 4032 & 1010 & 1319 & 80.0 & 75.4 & 77.6\\
\hspace{1em}Base (P+M) & 4126 & 948 & 1225 & 81.3 & 77.1 & 79.2\\
\hspace{1em}Large (P) & 3890 & 1441 & 1461 & 73.0 & 72.7 & 72.8\\
\hspace{1em}Large (P+M) & 3980 & 1456 & 1371 & 73.2 & 74.4 & 73.8\\
\bottomrule
\end{tabularx}
\vspace{1em}

\subsection{Relation extraction}
\noindent
\begin{tabularx}{.5\textwidth}{Xr@{\hspace{2ex}}r@{\hspace{2ex}}r@{\hspace{2ex}}r@{\hspace{2ex}}r@{\hspace{2ex}}r}
\toprule
DDI & TP & FP & FN & P & R & F1\\
\midrule
\cite{zhang2018drug} & - & - & - & 74.1 & 71.8 & 72.9\\
ELMo & - & - & - & 79.0 & 78.9 & 78.9\\
BioBERT & 786 & 229 & 193 & 77.4 & 80.3 & 78.8\\
Our BERT &  &  &  &  &  & \\
\hspace{1em}Base (P) & 737 & 172 & 242 & 81.1 & 75.3 & 78.1\\
\hspace{1em}Base (P+M) & 775 & 198 & 204 & 79.7 & 79.2 & 79.4\\
\hspace{1em}Large (P) & 788 & 206 & 191 & 79.3 & 80.5 & 79.9\\
\hspace{1em}Large (P+M) & 748 & 234 & 231 & 76.2 & 76.4 & 76.3\\
\bottomrule
\end{tabularx}
\vspace{1em}

\noindent
\begin{tabularx}{.5\textwidth}{Xr@{\hspace{1ex}}r@{\hspace{2ex}}r@{\hspace{2ex}}r@{\hspace{2ex}}r@{\hspace{2ex}}r}
\toprule
Chem-Prot & TP & FP & FN & P & R & F1\\
\midrule
\cite{peng2018extracting} & 1983  & 746  & 1475  & 72.7  & 57.4 & 64.1\\
ELMo & - & - & - & 66.7 & 66.6 & 66.6\\
BioBERT & 2359 & 803 & 1099 & 74.6 & 68.2 & 71.3\\
Our BERT &  &  &  &  &  & \\
\hspace{1em}Base (P) & 2443 & 834 & 1015 & 74.5 & 70.6 & 72.5\\
\hspace{1em}Base (P+M) & 2354 & 996 & 1104 & 70.3 & 68.1 & 69.2\\
\hspace{1em}Large (P) & 2610 & 948 & 848 & 73.4 & 75.5 & 74.4\\
\hspace{1em}Large (P+M) & 2355 & 1423 & 1103 & 62.3 & 68.1 & 65.1\\
\bottomrule
\end{tabularx}
\vspace{1em}

\noindent
\begin{tabularx}{.5\textwidth}{Xr@{\hspace{1ex}}r@{\hspace{2ex}}r@{\hspace{2ex}}r@{\hspace{2ex}}r@{\hspace{2ex}}r}
\toprule
i2b2 & TP & FP & FN & P & R & F1\\
\midrule
\cite{rink2011automatic} & - & - & - & 72.0 & 75.3 & 73.7\\
ELMo & - & - & - & 71.2 & 71.1 & 71.1\\
BioBERT & 4391 & 1474 & 1902 & 74.9 & 69.8 & 72.2\\
Our BERT &  &  &  &  &  & \\
\hspace{1em}Base (P) & 4592 & 1459 & 1701 & 75.9 & 73.0 & 74.4\\
\hspace{1em}Base (P+M) & 4683 & 1291 & 1610 & 78.4 & 74.4 & 76.4\\
\hspace{1em}Large (P) & 4684 & 1805 & 1609 & 72.2 & 74.4 & 73.3\\
\hspace{1em}Large (P+M) & 4700 & 1719 & 1593 & 73.2 & 74.7 & 73.9\\
\bottomrule
\end{tabularx}
\vspace{1em}

\subsection{Document classification}
\noindent
\begin{tabularx}{.5\textwidth}{Xr@{\hspace{2ex}}r@{\hspace{2ex}}r@{\hspace{2ex}}r@{\hspace{2ex}}r@{\hspace{2ex}}r}
\toprule
HoC & P & R & F1\\
\midrule
\cite{du2018ml} & 81.3 & 81.7 & 81.5\\
ELMo & 78.2 & 81.9 & 80.0\\
BioBERT & 83.4 & 82.4 & 82.9\\
Our BERT &  &  &  \\
\hspace{1em}Base (P) & 86.2 & 84.4 & 85.3\\
\hspace{1em}Base (P+M) & 84.0 & 82.3 & 83.1\\
\hspace{1em}Large (P) & 91.0 & 83.9 & 87.3\\
\hspace{1em}Large (P+M) & 88.8 & 82.1 & 85.3\\
\bottomrule
\end{tabularx}
\vspace{1em}
\end{document}